\documentclass[conference]{IEEEtran}
\IEEEoverridecommandlockouts
\usepackage{cite}
\usepackage{amsmath,amssymb,amsfonts}
\usepackage{algorithmic}
\usepackage{graphicx}
\usepackage{textcomp}
\usepackage{xcolor}
\usepackage{booktabs}
\usepackage{soul}
\usepackage{caption}
\usepackage{subcaption}
\usepackage{dialogue}
\usepackage{comment}
\usepackage{balance}

\usepackage[T1]{fontenc} 
\usepackage{setspace} 
\usepackage{dialogue}
\usepackage{calc} 

\newlength{\widestname}


\def\BibTeX{{\rm B\kern-.05em{\sc i\kern-.025em b}\kern-.08em
    T\kern-.1667em\lower.7ex\hbox{E}\kern-.125emX}}
\begin{document}

\title{Gaze Behavior During a Long-Term, In-Home, Social Robot Intervention for Children with ASD
}

\author{
    Rebecca Ramnauth$^{1}$, Frederick Shic$^{2, 3}$, Brian Scassellati$^{1}$ \\
    \IEEEauthorblockA{
        \small
        $^{1}$Department of Computer Science, Yale University, New Haven, CT, USA \\
        $^{2}$Center for Child Health, Behavior, and Development, Seattle Children’s Research Institute, Seattle, WA, USA \\
        $^{3}$Department of Pediatrics, University of Washington School of Medicine, Seattle, WA, USA \\
        {\tt\small rebecca.ramnauth@yale.edu}
    }
}
\maketitle
\begin{abstract}
Atypical gaze behavior is a diagnostic hallmark of Autism Spectrum Disorder (ASD), playing a substantial role in the social and communicative challenges that individuals with ASD face. This study explores the impacts of a month-long, in-home intervention designed to promote triadic interactions between a social robot, a child with ASD, and their caregiver. 
Our results indicate that the intervention successfully promoted appropriate gaze behavior, encouraging children with ASD to follow the robot's gaze, resulting in more frequent and prolonged instances of spontaneous eye contact and joint attention with their caregivers. Additionally, we observed specific timelines for behavioral variability and novelty effects among users. Furthermore, diagnostic measures for ASD emerged as strong predictors of gaze patterns for both caregivers and children. These results deepen our understanding of ASD gaze patterns and highlight the potential for clinical relevance of robot-assisted interventions.
\end{abstract}
\vspace{-10pt}
\vspace{-5pt}

\section{Introduction}
Social interactions involve complex exchanges of gaze. People rely on eye contact to direct attention to objects or events, respond to others' shift in attention \cite{driver1999gaze}, encourage prosocial behaviors \cite{tiitinen2014using, wirth2010eye}, and infer others' thoughts, desires, or intentions \cite{brooks2015connecting, von2014direct}. Recent findings emphasize the key role our gaze patterns play in coordinating joint activities \cite{sebanz2006joint} and facilitating social learning \cite{richardson2005looking, clark2004speaking, tomasello2007shared}. In essence, gaze serves a critical communicative function and its temporal dynamics provide valuable cues in a social exchanges.

Yet, eye contact in Autism Spectrum Disorders (ASD) is a subject of continuing discussion in the literature. Atypical gaze behavior is a diagnostic hallmark of ASD and contributes to many of the social and communicative challenges individuals with ASD face \cite{arnold2000eye, falck2015eye}. For example, individuals with ASD show a reduced motivation to share attention with others \cite{chevallier2012social}. As compared to neurotypicals, individuals with ASD initiate joint attention to a lesser extent, are less sensitive to social gaze, and tend to avoid eye contact \cite{del2021temporal, senju2009atypical}.

It is commonly believed that training appropriate gaze behavior will enhance one's overall social skills because it is considered a prerequisite for more complex behaviors \cite{krasny2003social}. Therefore, eye contact is often targeted first for ASD intervention \cite{arnold2000eye}. The intervention pedagogies are typically centered around positively reinforcing naturally-occurring incidences of eye contact \cite{hwang2000effects, rapp2019further}, modeling eye contact with others during social interactions \cite{weiss2010three}, or adjusting one's behavior by using visual supports to encourage eye contact with a speaker \cite{whalen2003joint, carbone2013teaching}. Although these interventions are intuitive methods for training appropriate gaze behavior, they demand the continued motivation of the caregiver, consistency in their behavioral feedback, and constant sensitivity to the specific needs and abilities of the individual with ASD over time. 

Socially assistive robotics (SARs) has the potential to augment the current efforts of caregivers and clinicians by eliciting positive and productive outcomes in ASD interventions \cite{scassellati2012robots}. The robots envisioned by these efforts support social and cognitive growth by improving access to on-demand, personalized, socially-situated, and physically co-present interventions. Research on SARs for ASD show increased engagement, improved attention regulation, and more appropriate social behavior such as joint attention and spontaneous imitation when robots are part of the interaction \cite{scassellati2012robots, pennisi2016autism}.

However, many of these studies focus on short-term interactions under controlled settings, or ultimately fail to demonstrate learning that generalizes to human-directed actions. In response to this critical gap in the literature, Scassellati et al. \cite{scassellati2018improving} reports directly assessed improvements in social skills in children with ASD following an in-home, month-long intervention conducted by an autonomous, socially assistive robot.
The study is a preliminary step to demonstrating that SARs are capable of producing lasting enhancements in social and communicative skills that are generalizable beyond the specific robot encounter to real-world, human-human interactions.

The rich dataset that resulted from this study characterized skills improvement using standard assessments at four time points: (i) 30 days before the intervention began; (ii) on the first day of the robot intervention; (iii) on the last day of the intervention; and (iv) 30 days after the end of the intervention. These assessments include measures of engagement based on the child's performance in various social skill games, joint attention between the child and their caregiver, and caregivers' surveys of their child's initiation of eye contact and verbal communication beyond the robot-assisted intervention.

Automated measures of performance in the dynamic, unconstrained environment that is the home, demand complex sensing. Therefore, the measures of gaze behavior reported by Scassellati et al. \cite{scassellati2018improving} were conducted manually by a clinician at these four time points. Although investigating improvement in this discretized fashion speaks broadly to the clinical effects of the training provided, an analysis of continuous change over the month-long intervention may better capture the subtle and nuanced behavioral patterns of ASD. However, in order to perform this analysis, a reliable method of gaze extraction must first be developed and then applied to the entire source dataset. The results produced by this automated method will, not only confirm those that were collected manually by a clinician, but also provide valuable insights into the sensing required to accurately detect gaze behavior in the home.  

Furthermore, the analysis examined only one aspect of gaze behavior, i.e., joint attention between the child and the caregiver. The source data contained substantial information about other forms of gaze behavior such as attentional shifts, mutual gaze, and gaze-following among the robot, child, caregiver, and other agents in the home environment. This additional exploration can provide a more comprehensive analysis of the effects of the intervention on gaze behavior in ASD.

In summary, Scassellati et al. \cite{scassellati2018improving} presented initial findings on the feasibility and efficacy of delivering an ASD intervention with a robot. This study expands the definitions, detection, and analysis of gaze behavior in \cite{scassellati2018improving} to better describe the effects of a long-term, in-home, social robot intervention on gaze behavior in ASD. The results of this expanded analysis have the potential to transform how we design and approach long-term, robot-assisted interventions for ASD. 

\section{Method}
Participants for the initial study were recruited through the university's medical school, and the current research team obtained Institutional Review Board approval to access their data. The following sections provide details on the participants, the design and components of the intervention system, and the methods used to extract and analyze gaze behavior from the interactions. Most of these details were not included in the initial study by Scassellati et al., thus supplementing the prior work and providing essential context for the current analysis.

\subsection{Participant Information}
Fifteen families with a child with ASD enrolled in the study. Two families withdrew, one due to unrelated health problems and one due to technical difficulties with the robot installation. Among the families who completed the study, five of the children were females and eight were males. The participants' age ranged from 6 to 12 years old ($M=10.0$, $SD=1.4$).

The diagnosis of ASD was established using a clinical best-estimate (CBE) approach by licensed psychologists and speech-language pathologists experienced in the diagnostic process. Prior to study inclusion, autism symptoms were characterized using the Autism Diagnostic Interview-Revised (ADI-R) and the Autism Diagnostic Observation Schedule (ADOS), gold-standard tools for clinical ASD diagnosis.

\begin{figure}[htbp]
  \includegraphics[width=\linewidth]{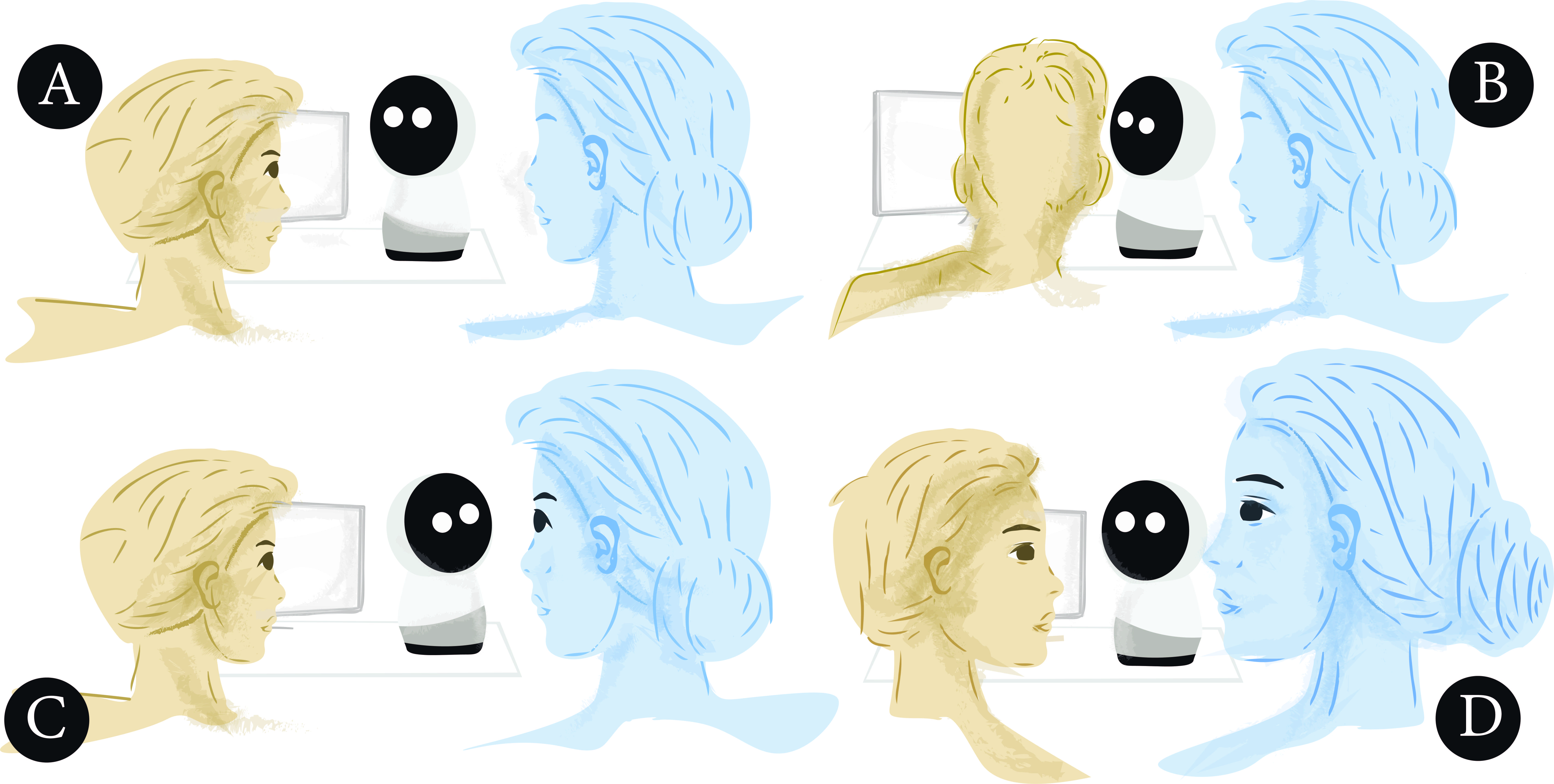}
  \setlength{\abovecaptionskip}{-10pt}
  \setlength{\belowcaptionskip}{-15pt}
  \caption{\textbf{Modeling Gaze}. The robot's context-contingent gaze guides the child's attention between the screen and caregiver, promoting increased interaction. See Sec. \ref{sec:content} for more details.} 
  \label{fig:sequence}
\end{figure}

The ADI-R is a semi-structured parent interview assessing four domains: reciprocal social interactions ($M=18.3$, $SD=6.9$, cutoff: 10), communication ($M=16.6$, $SD=4.7$, cutoff: 8), restricted and repetitive behaviors (RRB; $M=3.6$, $SD=0.8$, cutoff: 3), and early abnormal development history ($M=3.4$, $SD=0.7$, cutoff: 1). The ADOS is a clinician-administered behavioral assessment that provides a calibrated severity score ($M=7.3$, $SD=2.0$, cutoff: 4). While useful for symptom characterization, both tools were designed for diagnostic classification, not for measuring change, as they lack item granularity and sensitivity/specificity to short-term intervention outcomes \cite{anagnostou2015measuring, mcconachie2015systematic}. In this study, the ADI-R and ADOS were used to explore associations between detected changes and ASD symptomatology.

All participants had IQ scores $\geq 70$ as measured by the Differential Ability Scales (DAS-II), with means across verbal reasoning ($M=91.8$, $SD=25.9$), nonverbal reasoning ($M=95.2$, $SD=15.7$), spatial reasoning ($M=94.2$, $SD=16.0$), and general conceptual ability ($M=93.1$, $SD=19.6$). Participants exceeded ASD cutoffs on either the ADOS or ADI-R, alongside a confirmed CBE diagnosis.



\subsection{Robot-Assisted Intervention System}
We describe here the design and expected outcomes of the robot-assisted intervention, the content of the interactions, and the physical and technical components of the system. 

\textbf{Intervention Design}.
The robot-assisted intervention consisted of 30-minute sessions each day for 30 days and involved triadic opportunities for interaction and shared experiences among the robot, the child, and the caregiver. To achieve this, the robot was designed according to four primary goals: to (i) model realistic social behaviors; (ii) operate autonomously in the home; (iii) deliver personalized content; and (iv) facilitate interactions between the child with ASD and the caregiver. 



\textbf{Intervention Content}. \label{sec:content}
The intervention content consists of three interactive games that allowed children with ASD to practice social skills through play. Each of the three games targets one of three social skills: social and emotional understanding, perspective-taking, and ordering and sequencing. The design of these social games is further described in \cite{scassellati2018improving}. 


During the games, the robot demonstrates context-contingent gaze, as illustrated in Fig. \ref{fig:sequence}. When the child looks at the robot (A), the robot will direct the child's attention to the game content on the screen (B). After, the robot redirects the child's attention to the caregiver (C). We expect the child will follow the robot's gaze (D) and, thus, improve the frequency and duration of their interactions with their caregiver.

\begin{figure}[t]
  \includegraphics[width=\linewidth]{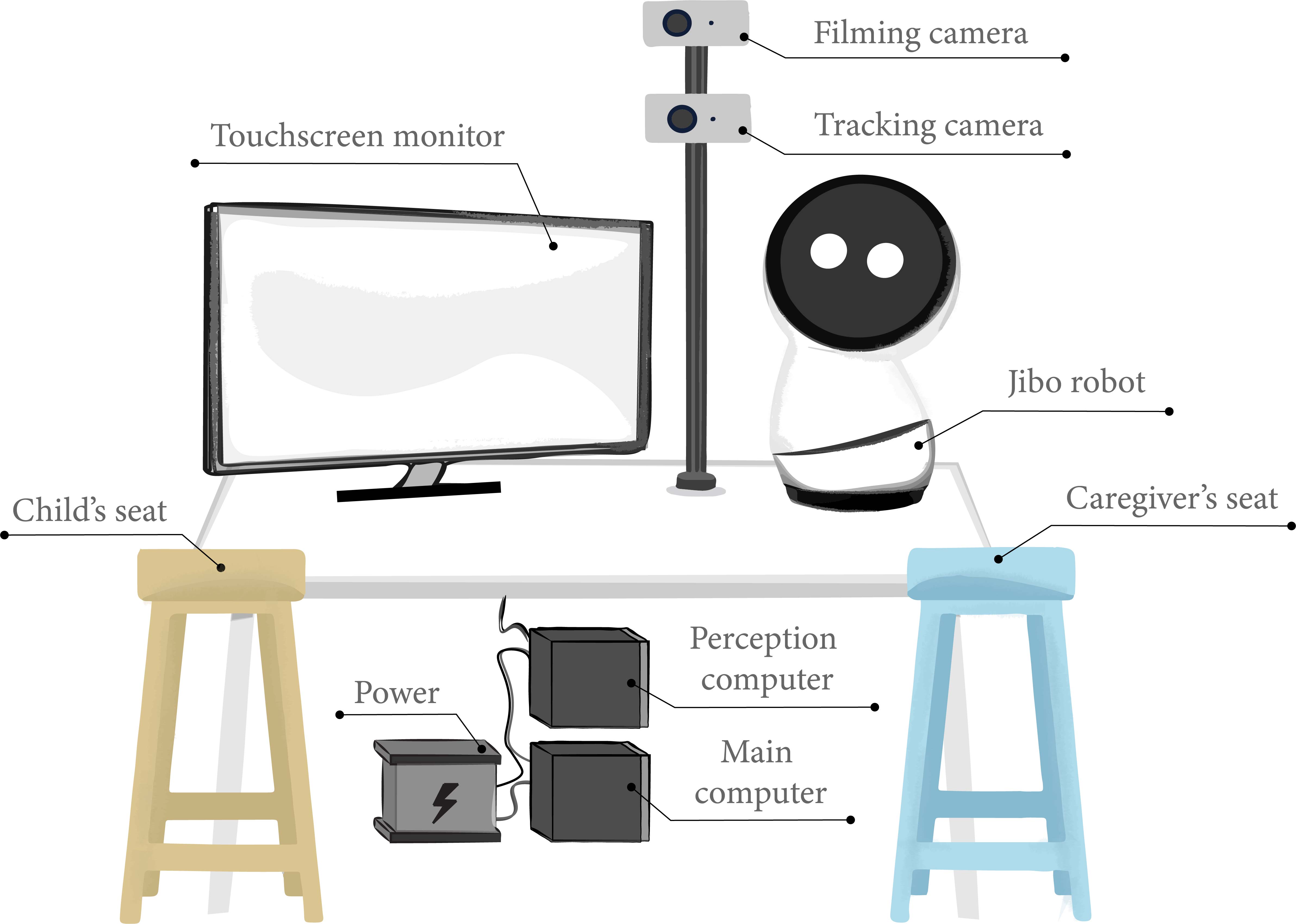}
  \setlength{\abovecaptionskip}{-8pt}
  \setlength{\belowcaptionskip}{-15pt}
  \caption{\textbf{System Hardware}. The system includes several components to coordinate the robot's behavior, content, and data collection during the intervention sessions.}
  \label{fig:setup}
\end{figure}

\textbf{System Components}.
We used the robot Jibo \cite{Jibo} which stands 11 inches tall and has 3 full-revolute axes designed for 360-degree movement. Jibo's onboard capabilities allowed for the programming of personified behaviors such as naturalistic gaze, pose, and movement. Other hardware included a touchscreen monitor, two RGB cameras, a perception computer, and a main computer. The setup is illustrated in Fig. \ref{fig:setup}.

Since the robot-assisted intervention relied on modeling appropriate gaze behavior using the robot, Jibo was developed to have a pair of animated eyes as opposed to its default single eye. 
The perception computer used an elevated camera to track users' attentional focus, relaying this data to the main computer to coordinate the robot's behavior and game content. The touchscreen monitor displayed game content and served as a shared medium between the robot, child, and caregiver. A second camera recorded the sessions for post-study analysis. All components operated within the ROS framework \cite{ros}. 



\subsection{Gaze Extraction}
A total of 156 hours of interaction was collected, with each child completing an average of 25 sessions over the month. 



\begin{figure}[t]
  \includegraphics[width=\linewidth]{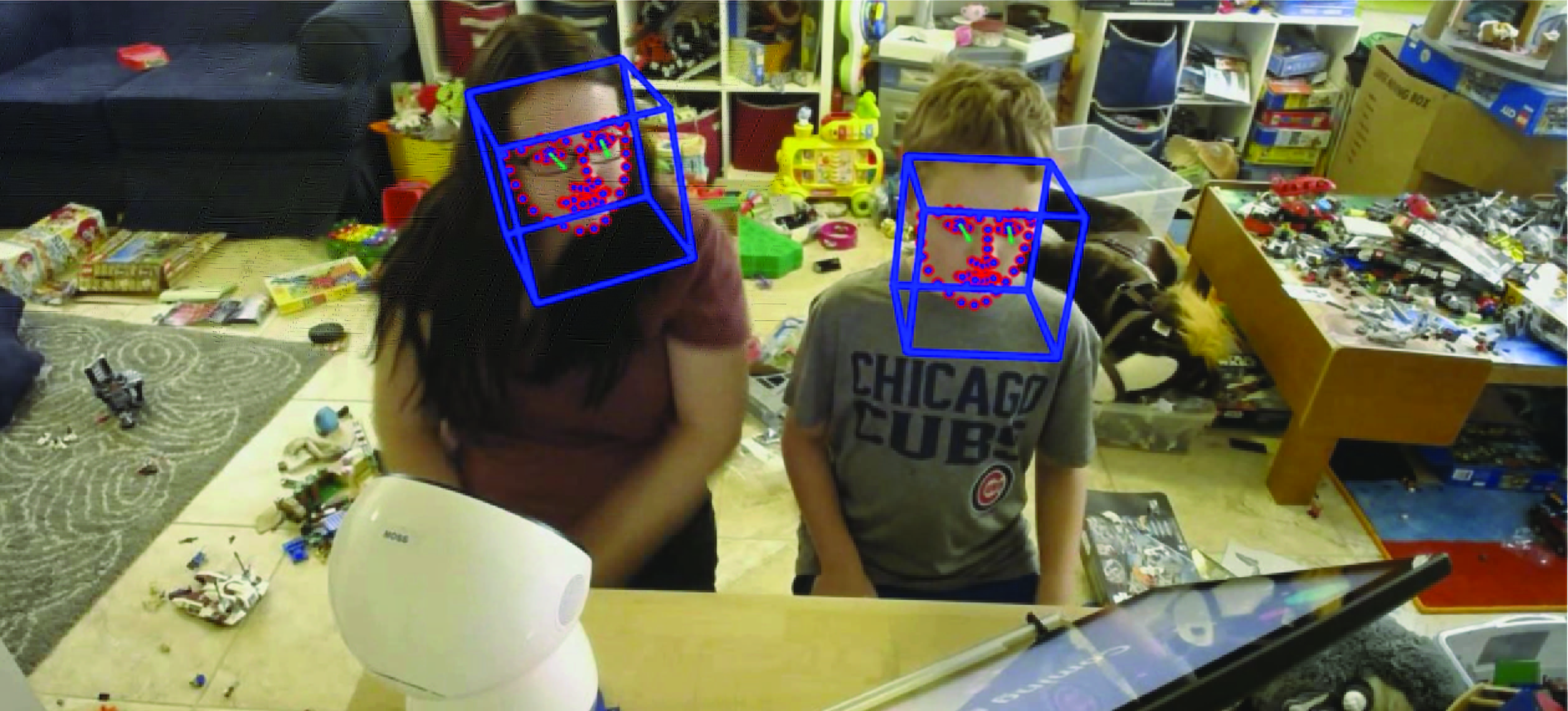}
  \setlength{\abovecaptionskip}{-8pt}
  \setlength{\belowcaptionskip}{-8pt}
  \caption{\textbf{Gaze Extraction}. We extract the several features such as gaze coordinates and facial landmarks to determine the gaze orientation of the child and their caregiver.}
  \label{fig:opencv}
\end{figure}

Each session recording was pre-processed using OpenFace \cite{baltrusaitis2018openface} to extract the gaze orientation of each person in the video feed (Fig. \ref{fig:opencv}). The resulting features represented information such as the gaze coordinates, facial landmarks, and facial action units for every image frame in the video recordings. Although the caregiver and child sat side-by-side during the intervention, we anticipated natural movement throughout the study, so their locations were not fixed in the analysis. Using OpenFace, we detected multiple faces in each video frame, designating the rightmost as the child and the leftmost as the caregiver. If more than two faces appeared—due to other people in the home—the correct faces were manually selected. 

To determine the attentional target of the participants, we defined the visual field of the child and caregiver as a cone. Since the location of static primary targets (i.e., screen and robot) are known and that of the caregiver relative to the child can be estimated, a person’s gaze is recorded when a target’s location falls within their visual cone. For each frame in each video recording, we extract the attentional targets of the child, caregiver, and robot as well as the start time and duration of attention on the target as measured in seconds. Targets include the robot, caregiver, child, screen, and unlabelled objects beyond the intervention content. Gaze data is compressed by grouping consecutive frames where attention remains on the same target, providing event-based data for shifts in attention. 

\textbf{Annotation Method}.
To assess the accuracy of the gaze detection algorithm, we performed annotations of the participant data. Since we are examining change over time, we randomly selected one session from the first two weeks and another from the last two weeks of each participant's study. 26 sessions were selected across the 13 participants for annotation, representing $12.4$ hours or $7.9\%$ of the total dataset. We used the ELAN software \cite{sloetjes2008annotation} to timestamp when the child, caregiver, or robot looked at a target and when they looked away.

\begin{table}[]
\caption{\textbf{Detection Accuracy}. The performances of the detection algorithm based on manual annotations are shown.}
\label{tab:accuracy}
\resizebox{\linewidth}{!}{%
\begin{tabular}{@{}llllllll@{}}
\toprule
\textbf{Gaze Component} & \textbf{N} & \textbf{Sensitivity} & \textbf{Specificity} & \textbf{PPV} & \textbf{NPV} & \textbf{AUC} & \textbf{F1} \\ \midrule
Individual Gaze     & 9,327  & 97\% & 93\% & 95\% & 90\% & 95\% & 96\% \\
Shared Gaze         & 6,972  & 96\% & 92\% & 94\% & 88\% & 93\% & 95\% \\
Mutual Gaze         & 5,823  & 93\% & 90\% & 92\% & 88\% & 92\% & 93\% \\
No Detection        & 1,195   & 91\% & 94\% & 92\% & 90\% & 93\% & 91\% \\ \midrule
Overall Performance & 23,317 & 94\% & 92\% & 93\% & 89\% & 94\% & 94\% \\ \bottomrule
\end{tabular}%
}
\vspace{-15pt}
\end{table}

To account for the fidelity of human transcriptions, the annotation representing the start of the gaze event was rounded down to the nearest quarter of a second and the annotation representing the end of the gaze event was rounded up to the nearest quarter of a second. As a result, $5,635$ total gaze events were annotated. We aligned all annotated events with the 23,317 detected by the algorithm based on video timestamps to measure overlap. The percentage overlap between detected and annotated events represents the algorithm's accuracy.

\textbf{Performance by Gaze Component}.
We identified three primary components of gaze behavior: (i) \emph{individual gaze}, where one person shifted attention to a target, (ii) \emph{shared gaze}, where two or more people looked at the same target, and (iii) \emph{mutual gaze}, where two people made eye contact. An event was labelled ``no detection'' when it could not classify, such as when one’s eyes were obscured or out of the camera’s view.

We evaluated the detection accuracy using several metrics, summarized in Table \ref{tab:accuracy}. Sensitivity was high for the three gaze types ($\geq$ 91\%). Positive Predictive Value (PPV), Negative Predictive Value (NPV), and the area under the ROC curve (AUC) confirmed the detections were sufficiently accurate. F1 scores provided a balanced view of precision and recall. Lastly, weighted averages summarized the overall performance.

\textbf{Performance by Subject}.
We assessed the performance of the detection algorithm in capturing the gaze behavior of both children and their caregivers. This evaluation provided essential insights into the algorithm's reliability across these two user groups in diverse home settings. We employed accuracy as the primary metric to assess whether algorithm correctly identified the attended target for each gaze event.

The detection algorithm demonstrated strong performance in characterizing the gaze behavior of caregivers ($M=94\%$, $SD=3.7\%$, $N=10,909$) and children ($M=88\%$, $SD=7.3\%$, $N=12,408$). Notably, the algorithm yielded a significant difference in average accuracy between caregiver and child data, as determined by a one-tailed z-test for sample proportions ($z=16.6$, $p\leq0.001$). This finding indicated that, on average, the algorithm more accurately detected the caregivers's gaze than the children's. Furthermore, an analysis of variance yielded a main effect of the individual, $F=24.7$, $p\leq0.001$, indicating that there is a significant difference in the algorithm's performance between caregivers and children.

In light of this, we investigated whether specific behaviors contributed to this accuracy difference. we observedd a significant difference for gaze events in which a caregiver looked at their child versus not ($94\%$, $N=1,522$, $z=3.7$, $p\leq0.001$) and a significant difference for gaze events in which a child looked at their caregiver versus not ($90\%$, $N=637$, $z=3.7$, $p\leq0.001$) regardless of whether the individual was engaging in independent gaze, shared attention, or mutual gaze. Altogether, this suggested that looking at a target to the immediate right or left significantly influences the accuracy of detection. This is to be expected as turning one's head decreases the amount of facial data to determine gaze. Yet, a significant majority ($57\%$, $z=28.2$, $p\leq0.001$) of the events that could not be categorized and received a label of ``no detection'' by the detection algorithm were of children data. We suspected that this may be because the children showed more physical movement throughout the study than did the caregivers, as confirmed by examining the session recordings. 


\begin{figure}[t]
  \includegraphics[width=\linewidth]{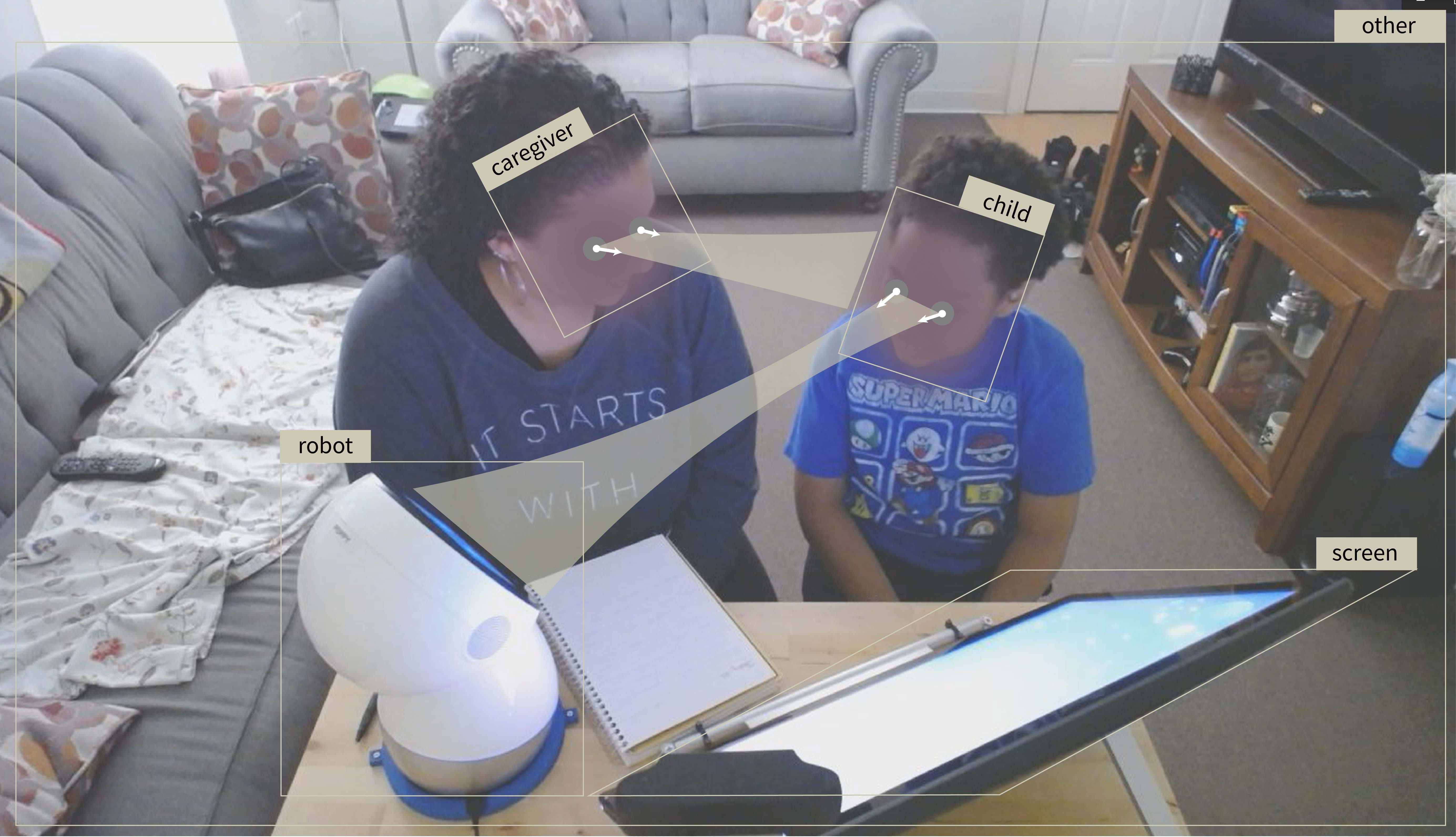}
  \setlength{\abovecaptionskip}{-8pt}
  \setlength{\belowcaptionskip}{-15pt}
  \caption{\textbf{Target Detection}. Attentional targets are estimated by the intersection of one's visual cone and static object locations.}
  \label{fig:visual-cone}
\end{figure}

 The difference in the algorithm's accuracy for gaze behavior between the caregiver and child is unsurprising, as most open-source datasets for automatic facial behavior analysis focus on neurotypical adults. The algorithm relies on OpenFace's eye gaze estimation \cite{baltruvsaitis2016openface}, which was evaluated by its authors using the MPIIGaze dataset \cite{zhang2015appearance}, collected from 15 adults during everyday laptop use. Despite growing interest in automatic gaze estimation, these methods have not been tested on individuals with ASD or children. Future research should investigate whether OpenFace and similar models are good surrogates for the behavioral annotation of these populations.  


Lastly, no significant variations in the algorithm's accuracy were observed when comparing the initial and final stages of the study, by week, or across other categories of gaze behavior.

\subsection{Dataset}

A total of $269,278$ gaze events resulted from this detection. This dataset thus describes gaze behaviors by the frequency and duration an individual engages in throughout their intervention. The current distribution of gaze duration is unimodal and positively skewed. 
Therefore, a log transform was applied to the duration of gaze to better conform the final dataset to normality, assessed using the Shapiro-Wilk test.

With the resulting dataset, we explore three main components of gaze behavior: (i) \emph{overall gaze} describing general attentional shifts to a target, (ii) \emph{mutual gaze} describing eye contact among the child, caregiver, and robot, and (iii) \emph{joint attention} between the child, caregiver, and robot in which two interacting partners first engage in eye contact, then one partner shifts their gaze to an object, causing their partner to orient their gaze to the same object. We describe these behaviors by the frequency and duration the child or caregiver engages in them over the course of the intervention. 

\section{Results}
For each component of gaze behavior, we calculated the averages and variances of gaze instances and duration. We also conducted multiple linear regression analyses to identify predictors of gaze instances for each attentional target. Similar models were used to determine predictors of gaze duration on each target and to assess the moderating effects of clinical measures, including the ADOS, ADI-R, and DAS-II.


We assessed the weekly effects on each gaze component while acknowledging that tasks varied day-to-day based on the children's interests and selections. Because the intervention system adapted to individual preferences, direct comparisons between children or per session were not feasible. However, at the weekly level, each participant was sufficiently exposed to the intervention, despite variations in the daily games and interaction content. Thus, we assessed behavioral changes across participants on a weekly basis throughout the intervention.

\subsection{Overall Gaze Behavior of the Child} 

We first investigate the children's distribution of attention across the various targets over time. A multiple linear regression was calculated to predict the log duration of the children’s gaze on each attentional target. The regression reveals a significant effect of the week ($F=19.5$, $p\leq0.001$) when the target is the caregiver ($\beta=-0.63$, $p \leq0.001$), when the target is the robot ($\beta=-0.29$, $p\leq0.001$), when the target is the screen ($\beta=0.34$, $p\leq0.001$), and when the target is other than these predefined targets ($\beta=-0.37$, $p\leq0.001$). Estimated coefficients are denoted as $\beta$. A regression did not reveal any significant effects of the clinical measures (ADOS, ADI-R, or DAS-II) on the children's distribution of gaze. 

A Tukey’s HSD reveals that the average duration of gaze occurrences with the robot ($M=6.31s$, $SD=1.08s$) and caregiver ($M=4.07s$, $SD=1.21s$) is significantly lower ($p\leq0.001$) than that with the screen ($M=70.8s$, $SD=2.66s$) or targets outside of the intervention ($M=13.2s$, $SD=1.40s$). 
Children's focus on the screen is expected, given that the game's content is a core part of the intervention. 
Thus, we further examine attention patterns for each target.

\subsubsection{Gaze with the Caregiver} \label{sec:childgaze-caregiver}
A paired t-test performed on the average number of attentional shifts toward the caregiver per week reveals a significant increase in the frequency a child shifts attention toward the caregiver between the first and the last week of the intervention ($t=3.38$, $p=0.005$).

A multiple linear regression calculated to predict the log duration of the child’s gaze on the caregiver revealed a significant effect of the week ($F=9.71$, $p\leq 0.001$). The post-hoc analysis reveals a significant decrease in gaze duration until the third week ($M=2.82s$, $SD=0.54s$, $\beta=-0.26$, $p\leq0.001$), where the first two weeks of the study resulted in a significant decrease in gaze duration ($\Delta M=-1.82s$, $p=0.003$) and a significant decrease in gaze duration between the second and third week ($\Delta M=-1.45s$, $p=0.006$). However, in the last week of the study, we observed a significant increase in the gaze duration of the child with the caregiver ($\Delta M=1.86s$, $p\leq0.001$). This change indicates that increased gaze duration with the caregiver occurred after having engaged with the intervention for at least three weeks. 


The multivariate linear regression showed no significant effect of clinical scores on children's gaze toward the caregiver. Overall, participants with ASD increased both their gaze frequency and duration toward the caregiver over time.


\subsubsection{Gaze with the Robot}
A paired t-test performed on the average number of attentional shifts towards the robot per week reveals a significant increase in the frequency a child with ASD shifts gaze to the robot beginning in the third week of the intervention ($t=2.65$, $p=0.03$). However, this significant increase does not persist into the last week of the intervention ($t=1.34$, $p=0.21$). This change indicates that participants with ASD showed an increased tendency of looking at the robot only after two weeks of the intervention.

A regression calculated to predict the log duration of the child’s gaze on the robot reveals a significant effect of the week ($F=18.8$, $p\leq 0.001$). The post-hoc analysis reveals a significant decrease in gaze duration with the robot throughout the study. The decrease becomes extremely significant in the third week of the study ($M=10.0s$, $SD=15.3s$, $\beta=-0.17$, $p\leq0.05$) as compared to the previous week ($M=14.5s$, $SD=5.14s$, $\beta=-0.07$). This rate of decrease persisted to the end of the study ($\beta=-0.17$, $p\leq0.001$) and therefore suggests that children with ASD were more likely to shift attention away from the robot over time.

However, a Levene variance test shows log gaze durations varied significantly by week ($w=7.96$, $p\leq0.001$). A significant change in the gaze duration variance in children with ASD occurred until two weeks into the study and persisted until the end ($p\leq0.001$). This change suggests that, although the rate of decreased attention to the robot was similar among all participants with ASD, the variability in gaze duration among participants with ASD was greater later in the intervention. 


In addition, the regression showed a significant effect of the ADOS severity score ($\beta=1.21$, $p=0.006$) and all categories of the ADI-R (reciprocal social interactions, $\beta=0.59$, $p=0.007$; communication, $\beta=-0.35$, $p=0.007$; restricted, repetitive, and stereotyped behaviors, $\beta=-0.87$, $p=0.005$; history of early abnormal development, $\beta=1.32$, $p=0.006$), and of the DAS-II (verbal reasoning, $\beta=0.02$, $p=0.02$; nonverbal reasoning $\beta=0.52$, $p=0.006$; spatial reasoning, $\beta=0.22$, $p=0.007$; GCASS, $\beta=-0.45$, $p=0.007$). Children with higher ASD severity scores, lower communicative ability, or more stereotyped behaviors were more likely to show increased attention toward the robot. 


\subsubsection{Gaze with the Screen}
A paired t-test performed on the number of attentional shifts reveals a significant increase in the frequency the children look at the screen between the first and last week of the intervention ($t=5.50$, $p\leq0.001$). 

A regression calculated to predict the log duration of the child’s gaze on the screen revealed a significant effect of the week ($F=76.8$, $p\leq 0.001$). The post-hoc analysis reveals a significant increase in gaze duration with the screen throughout the study ($\beta=0.33$, $p\leq0.001$), suggesting that children with ASD consistently attend longer to the screen over time.

However, 
a Levene variance test shows log gaze duration significantly varies among children with ASD by week ($w=24.43$, $p\leq0.001$). This significant change begins in the third week of the study ($M=1737.8s$, $SD=33.7s$), as compared to the previous week ($M=100.0s$, $SD=5.92s$), and persists to the end of the study. This suggests that, although the rate of increased attention to the screen is similar among all participants with ASD across each week, the variability in gaze duration is greater after two weeks into the intervention. 

The effect of clinical scores\footnote{ADOS calibrated severity score ($\beta = -1.04$, $p = 0.002$); all ADI-R categories: reciprocal social interactions ($\beta = -0.56$, $p = 0.001$), communication ($\beta = 0.36$, $p \leq 0.001$), restricted, repetitive, and stereotyped behaviors ($\beta = 0.76$, $p \leq 0.001$), history of early abnormal development ($\beta = -1.14$, $p = 0.002$); and all DAS-II categories: verbal reasoning ($\beta = 0.02$, $p = 0.01$), nonverbal reasoning ($\beta = -0.45$, $p = 0.002$), spatial reasoning ($\beta = -0.19$, $p = 0.002$), GCASS ($\beta = 0.39$, $p = 0.002$).} on gaze duration with the screen, although similar in magnitude, is in the opposite direction of that with the robot; children with lower severity and stereotyped behaviors, and higher communicative ability and IQ showed increasing attention toward the screen. 

\subsubsection{Gaze with Other Targets}
A paired t-test reveals a significant decrease in the frequency in which a participant with ASD turns attention to targets outside of the intervention between the first and last week ($t=4.32$, $p\leq0.001$). 


A multiple linear regression calculated to predict the log duration of the child’s gaze outside of the intervention's targets revealed a significant effect of the week ($F=7.06$, $p\leq0.001$). The post-hoc analysis reveals a significant increase in gaze duration with external targets after the second week ($\beta=0.05$, $p\leq 0.001$), but this significant increase is not observed throughout the study ($\beta=-0.01$, $p=0.82$). This transition after the second week ($M=10.72s$, $SD=12.26s$) is reflected in a significant increase in average gaze duration with external objects between the third week ($M=19.1s$, $SD=25.0s$, $p\leq0.001$), and significant decrease the fourth week ($M=7.08s$, $SD=4.37s$, $p=0.013$). The differences of variance between the weeks is also significant ($w=13.41$, $p=0.004$) after in the second week and supports previous results indicating that the behavioral variability among those with ASD was greater after two weeks into the intervention.

The effect of clinical scores\footnote{ADOS calibrated severity score ($\beta = 1.26$, $p \leq 0.001$); all ADI-R categories: reciprocal social interactions ($\beta = 0.66$, $p \leq 0.001$), communication ($\beta = -0.41$, $p \leq 0.001$), restricted, repetitive, and stereotyped behaviors ($\beta = -0.91$, $p \leq 0.001$), history of early abnormal development ($\beta = 1.39$, $p \leq 0.001$); and all DAS-II categories: verbal reasoning ($\beta = 0.01$, $p = 0.04$), nonverbal reasoning ($\beta = 0.55$, $p \leq 0.001$), spatial reasoning ($\beta = 0.23$, $p \leq 0.001$), GCASS ($\beta = -0.46$, $p \leq 0.001$).} on gaze duration is similar in both magnitude and direction as that with the robot; children with lower ASD severity scores, high communicative ability, or less stereotyped behaviors are more likely to show increased attention toward the robot. 

\begin{figure}[t]
  \includegraphics[width=\linewidth]{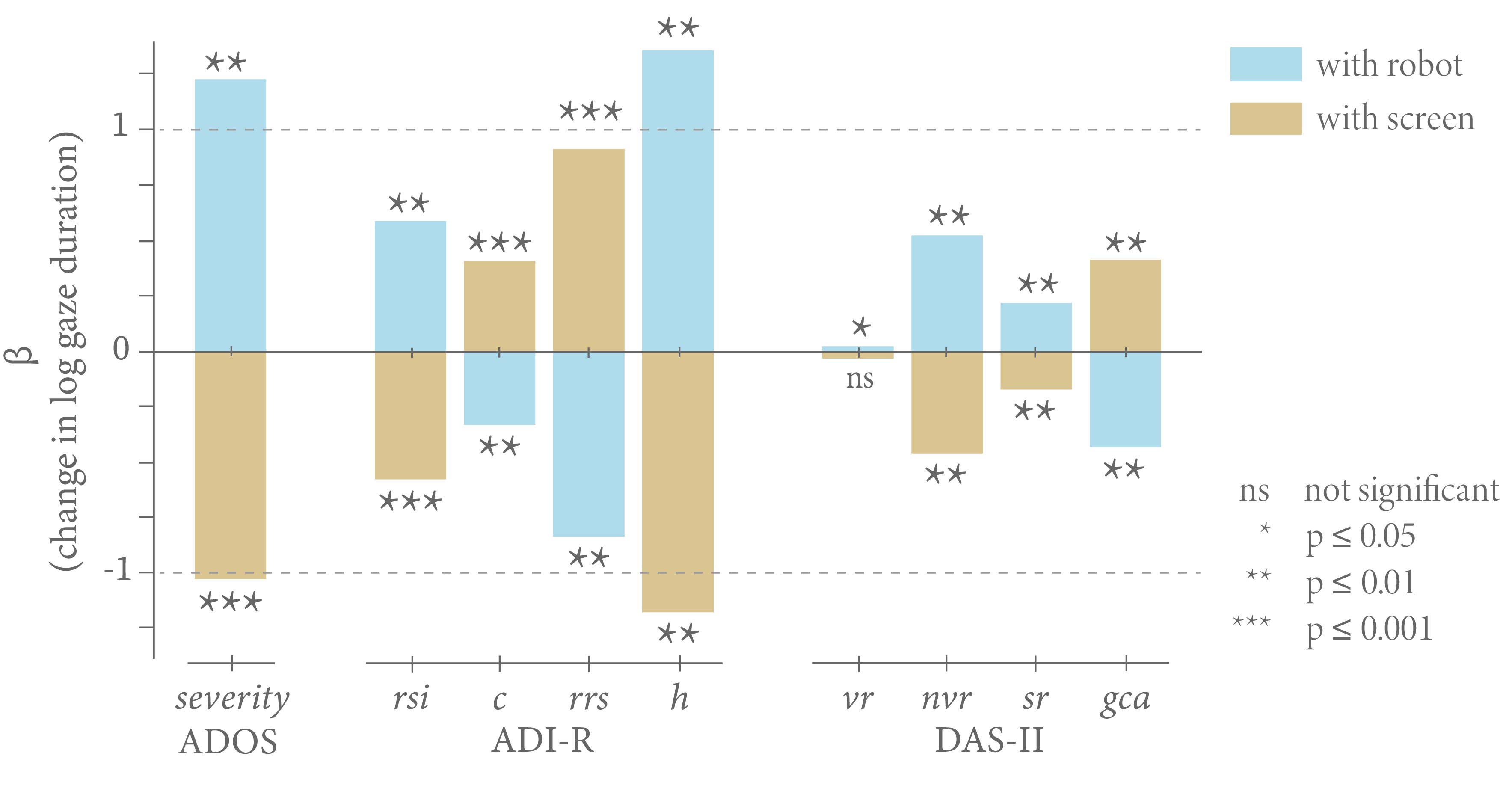}
  \setlength{\abovecaptionskip}{-12pt}
  \setlength{\belowcaptionskip}{-15pt}
  \caption{\textbf{Clinical Severity on Gaze}. The effect of ASD severity on gaze behavior toward the robot is similar in magnitude to, yet of the opposite direction of that with the screen.}
  \label{fig:clinical-duration}
\end{figure}

\subsection{Overall Gaze Behavior of the Caregiver}
Using a paired t-test, we found a significant increase in caregiver gaze shifts to the robot from the first to the last week of the intervention ($t=7.97$, $p\leq0.001$), with no significant change in gaze to the screen ($t=1.33$, $p=0.21$). Caregivers showed a significant decrease in gaze toward the child over the study period ($t=-15.2$, $p\leq0.001$) and a significant increase in shifts to external targets ($t=-3.82$, $p=0.002$).

\begin{figure}[t]
  \includegraphics[width=\linewidth]{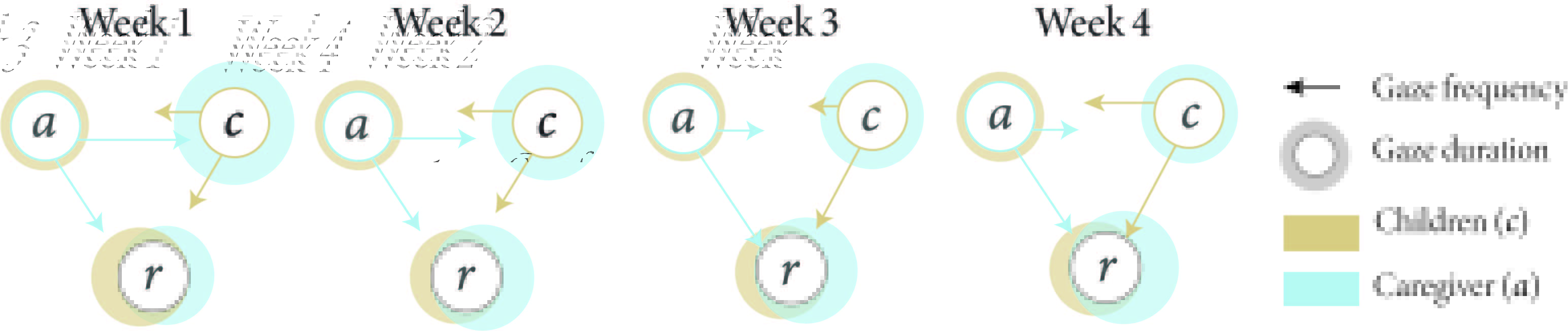}
  \setlength{\abovecaptionskip}{-8pt}
  \setlength{\belowcaptionskip}{-15pt}
  \caption{\textbf{Change by Week}. Average gaze duration and frequency for adult caregivers (\emph{a}) and children (\emph{c}) are shown. Circle diameters indicate average duration (in seconds) toward each other and robot (\emph{r}), while line lengths indicate frequency. Bar chart representations are provided in the Appendix.}
  \label{fig:gaze-change}
\end{figure}

A regression of the log duration of the caregiver’s gaze also reveals a significant effect of the week ($F=70.7$, $p\leq0.001$) and when the target is the child ($\beta=-0.23$, $p\leq0.001$), robot ($\beta=0.16$, $p\leq0.001$), screen ($\beta=0.67$, $p\leq0.001$), or outside of these predefined targets ($\beta=0.22$, $p\leq0.001$).

A Tukey HSD revealed that caregivers spent significantly more time attending to the screen ($M=166.0s$, $SD=3.95s$) compared to other targets, including those outside the interaction ($M=32.4s$, $SD=3.36s$), the robot ($M=14.1s$, $SD=2.32s$, $p=0.03$), and their child ($M=6.03s$, $SD=1.32s$, $p\leq0.001$). The increased attention to the robot, screen, and other targets beyond the intervention indicates a quicker shift in focus away from the child. 
Although children's gaze duration and frequency toward their caregiver increased significantly (though not consistently across the weeks, see Sec. \ref{sec:childgaze-caregiver}), caregivers’ gaze toward their child decreased in a more pronounced, consistent manner over time. Fig. \ref{fig:gaze-change} shows the magnitude of change for the children and caregivers.


The regression calculated to predict the log duration of the caregiver’s overall gaze reveals significant effects of their child’s ADOS calibrated severity score ($\beta=1.00$, $p\leq0.001$), all categories of the ADI-R (reciprocal social interactions, $\beta=0.46$, $p\leq0.001$; communication, $\beta=-0.28$, $p\leq0.001$; restricted, repetitive, and stereotyped behaviors, $\beta=-0.73$, $p\leq0.001$; history of early abnormal development, $\beta=1.06$, $p\leq0.001$), and all categories of the DAS-II (verbal reasoning, $\beta=0.02$, $p\leq0.001$; nonverbal reasoning, $\beta= 0.42$, $p\leq0.001$; spatial reasoning, $\beta=0.19$, $p\leq0.001$, GCASS, $\beta=-0.37$, $p\leq0.001$). Further investigation reveals that the significant effects of clinical measures occur only when caregivers focused on the robot or child. This suggests that caregivers of children with high ASD severity scores engaged in longer gaze behavior with both the robot and child over time. Moreover, the direction of these effects is similar for both caregivers and children: both engage in longer gaze with the robot when clinical measures indicate high ASD severity, low communicative ability, or more stereotyped behaviors.

\subsection{Joint Attention Based on Mutual Gaze}

The previous analyses focused on trends in individual gaze instances and durations, as well as predictors such as weekly exposure to the intervention and clinical measures. We now expand our scope to examine the contingency of gaze among the robot, children, and caregivers throughout the intervention. We define this contingency through \emph{joint attention involving eye contact}, which occurs when two individuals engage in mutual gaze, and one shifts their gaze to an object, prompting the other to follow. This gaze following reflects an expectation-based orienting, where one person's change in gaze cues the other's attention \cite{macpherson2017attentional}. It is anticipated that joint attention initiated by mutual gaze leads to greater motivation to follow gaze cues and longer durations of shared gaze \cite{vernon2014fostering, mundy1997joint}.

\subsubsection{Between Child and Caregiver}
A paired t-test revealed a significant increase in spontaneous mutual gaze between children and caregivers ($t=4.31$, $p=0.009$). A regression predicting the duration of shared gaze following joint attention showed a significant effect of the week ($F=10.30$, $p\leq0.001$), with a marked increase in the first week that persisted throughout the study ($\beta=0.27$, $p\leq0.001$). These findings suggest that joint attention between children and caregivers increased, with more frequent mutual gaze leading to longer periods of shared gaze over time. No significant effects of clinical measures were found in the regression.

We also observed a significant decrease in the duration of mutual gaze from the first to the last week of the study ($\Delta M=-1.43s$, $\beta=-0.20$, $p=0.002$). Despite the significant increase in joint attention and the resulting shared gaze between the child and caregiver, the duration of mutual gaze decreased. This may be viewed as a positive outcome, as shorter durations of the joint attention cue (i.e., eye contact) allows for longer durations of shared attention. 


\subsubsection{Between Robot and Child}
A paired t-test on the frequency of the child's gaze toward their caregiver following eye contact with the robot indicates a significant increase after the second week of the intervention ($t = 4.56$, $p \leq 0.001$). Similarly, after the second week, when the child shifted gaze away from the caregiver, they more frequently redirected their attention back to the robot ($t = 3.46$, $p = 0.004$).
A regression analysis also shows a significant effect of the week ($F = 8.44$, $p \leq 0.001$). Joint attention between the robot and child significantly increased from the first week and continues through the study ($\beta = 0.03$, $p = 0.006$). 


We observed a similar trend in the duration of eye contact between the child and the robot ($\beta = 0.05$, $p \leq 0.001$) and in gaze duration when the joint attentional target is the caregiver ($\beta = 0.27$, $p = 0.05$). The increase in joint attention indicates that the children directed more attention to their caregiver while engaging with the robot over time. The regression, however, does not show a significant effect of clinical scores.

\subsubsection{Between Robot and Caregiver}
A paired t-test of the frequency of gaze of the caregiver towards their child following eye contact between the robot and the caregiver by week suggests that the caregiver significantly shifted gaze more often toward the child after following the gaze of the robot throughout the study ($t=2.80$, $p=0.02$). When shifting gaze away from the child, caregivers shifted their attention more often to the screen ($t=10.1$, $p=0.004$).

A regression predicting instances of joint attention between the caregiver and robot shows a significant weekly effect ($F = 5.00$, $p = 0.002$), with a notable increase starting after the second week and persisting throughout the intervention ($\beta = 0.07$, $p = 0.004$). Additionally, we observed a significant increase in the duration of eye contact between the caregiver and robot ($\beta = 0.33$, $p \leq 0.001$), as well as in gaze duration when the child is the joint attentional target ($\beta = 0.07$, $p = 0.006$), but not when the target is the screen ($\beta = -0.05$, $p = 0.52$). This suggests that caregivers increasingly focused on their children while engaging with the robot over time.

The effects of a child's clinical severity on their caregiver's overall gaze\footnote{ADOS calibrated severity score ($\beta = 1.38$, $p \leq 0.001$); ADI-R categories: reciprocal social interactions ($\beta = 0.70$, $p \leq 0.001$); communication ($\beta = -0.43$, $p \leq 0.001$); restricted, repetitive, and stereotyped behaviors ($\beta = -0.99$, $p \leq 0.001$); history of early abnormal development ($\beta = 1.49$, $p \leq 0.001$); DAS-II categories: verbal reasoning ($\beta = 0.03$, $p \leq 0.001$); nonverbal reasoning ($\beta = 0.61$, $p \leq 0.001$); spatial reasoning ($\beta = 0.26$, $p \leq 0.001$); GCASS ($\beta = -0.53$, $p \leq 0.001$).} and the joint attention between the caregiver and robot are similar in both direction and magnitude; joint attention between the caregiver and robot increased when their child exhibits higher ASD severity, lower communicative ability, or more stereotyped behaviors.

\section{Discussion}
Scassellati et al. \cite{scassellati2018improving} introduced a robot-assisted intervention system that provided personalized, on-demand cognitive and social support for children with ASD. We expand the definitions, detection methods, and analysis of user behavior to better capture the effects of a long-term, in-home social robot intervention for ASD. Our findings center on three key themes: (i) the intervention improved gaze behavior in children with ASD; (ii) behavioral variability among participants increased significantly after two weeks of engagement; and (iii) diagnostic measures like the ADI-R, ADOS, and DAS-II proved to be strong predictors of behavioral change for both caregivers and children. These insights are crucial for designing effective robot-assisted social skills interventions and understanding behavioral trends in ASD.

\subsection{Improvements in Gaze Behavior}
The social robot promoted appropriate gaze behavior during the intervention, leading to improved spontaneous gaze between children with ASD and their caregivers. Children were significantly more likely to direct their attention and make eye contact with their caregivers. Our analysis revealed significant increases in instances of joint attention, spontaneous mutual gaze, and the duration of shared gaze between the pairs. However, while children engaged in eye contact with their caregivers more frequently, the duration of eye contact prior to shared gaze decreased over time. We view this as a positive outcome, as it indicates that children needed less time for the joint attentional cue (eye contact) to maintain longer periods of shared attention with their caregivers.

Furthermore, the children's gaze with the caregiver was contingent on that of the robot throughout the study: children with ASD were more likely to engage in longer eye contact with their caregiver after they saw the robot shift its attention to the caregiver. This contingency of gaze is also true of caregivers: caregivers were more likely to engage in longer eye contact with their child after they saw the robot shift its attention to the child. This suggests that a robot designed to redirect a person's attention by modeling the shift in gaze may be effective at improving the frequency of eye contact.

Ultimately, gaze following with the robot was natural, increased throughout the study for both children and caregivers, and encouraged more frequent eye contact and shared attention between the children and caregivers. Using a joint attention probe, Scassellati et al. \cite{scassellati2018improving} found significant improvements in joint attention among children with ASD following a robot intervention. Similarly, this study confirms consistent joint attention gains with both the caregiver and the robot. Yet, we must acknowledge that the impact of the robot or any other system component cannot be measured independently. Furthermore, the sustainability of the observed gains may depend on ongoing participation in the intervention or additional support. These improvements were noted during the intervention, but further research is necessary to determine whether they persist long after the study concludes. We present this as a limitation of this study and an area for future work.



\subsection{Timing \& Variability of Skill Improvements}
Several improvements in gaze behavior emerged \emph{only} until two weeks into the intervention. For instance, joint attention between the robot and caregivers significantly increased only after the second week. Additionally, children's gaze duration toward their caregiver initially decreased during the first two weeks, followed by a significant increase in the final two weeks. Hence, we recommend that similar social skills interventions be evaluated over a duration longer than two weeks to better capture the potential for significant behavioral change. 

It is well known that individuals with ASD show a broad spectrum of challenges and (dis)abilities, and vary greatly in their levels of social functioning. 
Although the participants were high-functioning individuals with ASD and able to understand the intervention's content, we observed significant variability in the gaze behaviors between users \emph{only} after two weeks. This variability among the children was especially evident in gaze towards objects that were initially novel: the screen and the robot. While each child's gaze behavior with these objects followed a similar pattern for the first two weeks, their behaviors with these objects diverged significantly after. Based on these findings, we recommend that interventions aimed at improving gaze behaviors in children with ASD be evaluated for more than two weeks, allowing for novelty effects to subside and increased individual variability to emerge. 

\subsection{Predictive Power of Diagnostic Measures}

Scassellati et al.’s joint attention probe found that children with lower nonverbal ability, as measured by the DAS-II, showed greater gains in joint attention skills. Our analysis further supports this, revealing a strong positive relationship between nonverbal ability and joint attention, suggesting that children with lower nonverbal reasoning had more capacity to grow in terms of joint attention skills. 

Furthermore, the strong correlation between clinical measures of ASD severity and gaze behaviors suggests that these metrics can be valuable for predicting intervention outcomes. For example, children with higher ASD severity and lower nonverbal ability showed increased attention to the robot while their attention to the screen decreased over time. These scores not only predicted the children's behaviors but also their caregivers’. Caregivers of children with high ASD severity engaged in longer gaze interactions with both the robot and their child. Being able to anticipate user outcomes based on clinical severity can influence how we think about the intervention's effectiveness and allow researchers to streamline the process by reducing the need for constant clinician oversight.

\section{Conclusion}
This study analyzes gaze behavior in a long-term, in-home social robot intervention for ASD. Our findings contribute several recommendations for designing social skills interventions and for understanding gaze behavior in ASD. The primary findings demonstrate that the robot-assisted intervention improved various aspects of gaze behavior in children with ASD, that there is a marked difference in when the children improve, and diagnostic measures can be good predictors of long-term gaze behavior of both caregivers and children with ASD.  


We recommend that designers of social skills interventions for ASD leverage these findings by recognizing the potential of robots to foster appropriate gaze behavior among users. Additionally, the results indicate that such an intervention on gaze behavior should be evaluated for at least two weeks to account for the decline of novelty effects and the subsequent behavioral variability among individual users. Further research is necessary to determine how effectively clinical assessments of ASD predict the outcomes of robot-assisted social skills interventions. The strong correlation between clinical scores and gaze behavior suggests that these assessments could reliably predict the behaviors of both children with ASD and their caregivers during the intervention. This relationship may reduce the need for constant oversight by a clinician or for disrupting in-home interactions to administer tests that may not fully capture the specific skills targeted by the intervention.



While our study provides valuable insights into gaze behavior during a long-term, in-home social robot intervention for ASD, it has several limitations. The small sample size (13 children with ASD and 13 caregivers) limits the generalizability of our findings. Behavioral changes from training often require weeks or months, and while the month-long intervention captured novelty effects and early impacts, longer studies are needed to assess the sustainability of improvements. Future research with larger samples and extended interventions is essential to better understand the long-term effects of social robot interventions on gaze behavior in ASD.


\section*{Acknowledgments}
\noindent This work was funded by the National Science Foundation (NSF) awards IIS-2106690, IIS-1955653, NSF Expedition 1139078, the Office of Naval Research (ONR) award N00014-24-1-2124, and the National Institutes of Health (NIH) K01 MH104739. R. Ramnauth is supported by both the NSF GRFP and the National Academics (NASEM) Ford Fellowships.

\bibliographystyle{IEEEtran}

\begin{thebibliography}{10}
\providecommand{\url}[1]{#1}
\csname url@samestyle\endcsname
\providecommand{\newblock}{\relax}
\providecommand{\bibinfo}[2]{#2}
\providecommand{\BIBentrySTDinterwordspacing}{\spaceskip=0pt\relax}
\providecommand{\BIBentryALTinterwordstretchfactor}{4}
\providecommand{\BIBentryALTinterwordspacing}{\spaceskip=\fontdimen2\font plus
\BIBentryALTinterwordstretchfactor\fontdimen3\font minus \fontdimen4\font\relax}
\providecommand{\BIBforeignlanguage}[2]{{%
\expandafter\ifx\csname l@#1\endcsname\relax
\typeout{** WARNING: IEEEtran.bst: No hyphenation pattern has been}%
\typeout{** loaded for the language `#1'. Using the pattern for}%
\typeout{** the default language instead.}%
\else
\language=\csname l@#1\endcsname
\fi
#2}}
\providecommand{\BIBdecl}{\relax}
\BIBdecl

\bibitem{driver1999gaze}
J.~Driver~IV, G.~Davis, P.~Ricciardelli, P.~Kidd, E.~Maxwell, and S.~Baron-Cohen, ``Gaze perception triggers reflexive visuospatial orienting,'' \emph{Visual cognition}, vol.~6, no.~5, pp. 509--540, 1999.

\bibitem{tiitinen2014using}
S.~Tiitinen and J.~Ruusuvuori, ``Using formulations and gaze to encourage parents to talk about their and their children's health and well-being,'' \emph{Research on Language and Social Interaction}, vol.~47, no.~1, pp. 49--68, 2014.

\bibitem{wirth2010eye}
J.~H. Wirth, D.~F. Sacco, K.~Hugenberg, and K.~D. Williams, ``Eye gaze as relational evaluation: Averted eye gaze leads to feelings of ostracism and relational devaluation,'' \emph{Personality and social psychology bulletin}, vol.~36, no.~7, pp. 869--882, 2010.

\bibitem{brooks2015connecting}
R.~Brooks and A.~N. Meltzoff, ``Connecting the dots from infancy to childhood: A longitudinal study connecting gaze following, language, and explicit theory of mind,'' \emph{Journal of experimental child psychology}, vol. 130, pp. 67--78, 2015.

\bibitem{von2014direct}
E.~A. von~dem Hagen, R.~S. Stoyanova, J.~B. Rowe, S.~Baron-Cohen, and A.~J. Calder, ``Direct gaze elicits atypical activation of the theory-of-mind network in autism spectrum conditions,'' \emph{Cerebral cortex}, vol.~24, no.~6, pp. 1485--1492, 2014.

\bibitem{sebanz2006joint}
N.~Sebanz, H.~Bekkering, and G.~Knoblich, ``Joint action: bodies and minds moving together,'' \emph{Trends in cognitive sciences}, vol.~10, no.~2, pp. 70--76, 2006.

\bibitem{richardson2005looking}
D.~C. Richardson and R.~Dale, ``Looking to understand: The coupling between speakers' and listeners' eye movements and its relationship to discourse comprehension,'' \emph{Cognitive science}, vol.~29, no.~6, pp. 1045--1060, 2005.

\bibitem{clark2004speaking}
H.~H. Clark and M.~A. Krych, ``Speaking while monitoring addressees for understanding,'' \emph{Journal of memory and language}, vol.~50, no.~1, pp. 62--81, 2004.

\bibitem{tomasello2007shared}
M.~Tomasello and M.~Carpenter, ``Shared intentionality,'' \emph{Developmental science}, vol.~10, no.~1, pp. 121--125, 2007.

\bibitem{arnold2000eye}
A.~Arnold, R.~J. Semple, I.~Beale, and C.~M. Fletcher-Flinn, ``Eye contact in children's social interactions: What is normal behaviour?'' \emph{Journal of Intellectual and Developmental Disability}, vol.~25, no.~3, pp. 207--216, 2000.

\bibitem{falck2015eye}
T.~Falck-Ytter, C.~Carlstr{\"o}m, and M.~Johansson, ``Eye contact modulates cognitive processing differently in children with autism,'' \emph{Child development}, vol.~86, no.~1, pp. 37--47, 2015.

\bibitem{chevallier2012social}
C.~Chevallier, G.~Kohls, V.~Troiani, E.~S. Brodkin, and R.~T. Schultz, ``The social motivation theory of autism,'' \emph{Trends in cognitive sciences}, vol.~16, no.~4, pp. 231--239, 2012.

\bibitem{del2021temporal}
T.~Del~Bianco, L.~Mason, T.~Charman, J.~Tillman, E.~Loth, H.~Hayward, F.~Shic, J.~Buitelaar, M.~H. Johnson, E.~J. Jones \emph{et~al.}, ``Temporal profiles of social attention are different across development in autistic and neurotypical people,'' \emph{Biological Psychiatry: Cognitive Neuroscience and Neuroimaging}, vol.~6, no.~8, pp. 813--824, 2021.

\bibitem{senju2009atypical}
A.~Senju and M.~H. Johnson, ``Atypical eye contact in autism: models, mechanisms and development,'' \emph{Neuroscience \& Biobehavioral Reviews}, vol.~33, no.~8, pp. 1204--1214, 2009.

\bibitem{krasny2003social}
L.~Krasny, B.~J. Williams, S.~Provencal, and S.~Ozonoff, ``Social skills interventions for the autism spectrum: Essential ingredients and a model curriculum,'' \emph{Child and Adolescent Psychiatric Clinics}, vol.~12, no.~1, pp. 107--122, 2003.

\bibitem{hwang2000effects}
B.~Hwang and C.~Hughes, ``The effects of social interactive training on early social communicative skills of children with autism,'' \emph{Journal of autism and developmental disorders}, vol.~30, no.~4, pp. 331--343, 2000.

\bibitem{rapp2019further}
J.~T. Rapp, J.~L. Cook, R.~Nuta, C.~Balagot, K.~Crouchman, C.~Jenkins, S.~Karim, and C.~Watters-Wybrow, ``Further evaluation of a practitioner model for increasing eye contact in children with autism,'' \emph{Behavior Modification}, vol.~43, no.~3, pp. 389--412, 2019.

\bibitem{weiss2010three}
M.~J. Weiss and T.~Zane, ``Three important things to consider when starting intervention for a child diagnosed with autism,'' \emph{Behavior Analysis in Practice}, vol.~3, no.~2, p.~58, 2010.

\bibitem{whalen2003joint}
C.~Whalen and L.~Schreibman, ``Joint attention training for children with autism using behavior modification procedures,'' \emph{Journal of Child psychology and psychiatry}, vol.~44, no.~3, pp. 456--468, 2003.

\bibitem{carbone2013teaching}
V.~J. Carbone, L.~O'Brien, E.~J. Sweeney-Kerwin, and K.~M. Albert, ``Teaching eye contact to children with autism: A conceptual analysis and single case study,'' \emph{Education and treatment of children}, vol.~36, no.~2, pp. 139--159, 2013.

\bibitem{scassellati2012robots}
B.~Scassellati, H.~Admoni, and M.~Matari{\'c}, ``Robots for use in autism research,'' \emph{Annual review of biomedical engineering}, vol.~14, pp. 275--294, 2012.

\bibitem{pennisi2016autism}
P.~Pennisi, A.~Tonacci, G.~Tartarisco, L.~Billeci, L.~Ruta, S.~Gangemi, and G.~Pioggia, ``Autism and social robotics: A systematic review,'' \emph{Autism Research}, vol.~9, no.~2, pp. 165--183, 2016.

\bibitem{scassellati2018improving}
B.~Scassellati, L.~Boccanfuso, C.-M. Huang, M.~Mademtzi, M.~Qin, N.~Salomons, P.~Ventola, and F.~Shic, ``Improving social skills in children with asd using a long-term, in-home social robot,'' \emph{Science Robotics}, vol.~3, no.~21, p. eaat7544, 2018.

\bibitem{anagnostou2015measuring}
E.~Anagnostou, N.~Jones, M.~Huerta, A.~K. Halladay, P.~Wang, L.~Scahill, J.~P. Horrigan, C.~Kasari, C.~Lord, D.~Choi \emph{et~al.}, ``Measuring social communication behaviors as a treatment endpoint in individuals with autism spectrum disorder,'' \emph{Autism}, vol.~19, no.~5, pp. 622--636, 2015.

\bibitem{mcconachie2015systematic}
H.~McConachie, J.~R. Parr, M.~Glod, J.~Hanratty, N.~Livingstone, I.~P. Oono, S.~Robalino, G.~Baird, B.~Beresford, T.~Charman \emph{et~al.}, ``Systematic review of tools to measure outcomes for young children with autism spectrum disorder,'' 2015.

\bibitem{Jibo}
\BIBentryALTinterwordspacing
N.~Disruption. (2020) Jibo. [Online]. Available: \url{https://jibo.com/}
\BIBentrySTDinterwordspacing

\bibitem{ros}
\BIBentryALTinterwordspacing
{Stanford Artificial Intelligence Laboratory et al.}, ``Robotic operating system.'' [Online]. Available: \url{https://www.ros.org}
\BIBentrySTDinterwordspacing

\bibitem{baltrusaitis2018openface}
T.~Baltrusaitis, A.~Zadeh, Y.~C. Lim, and L.-P. Morency, ``Openface 2.0: Facial behavior analysis toolkit,'' in \emph{2018 13th IEEE international conference on automatic face \& gesture recognition (FG 2018)}.\hskip 1em plus 0.5em minus 0.4em\relax IEEE, 2018, pp. 59--66.

\bibitem{sloetjes2008annotation}
\BIBentryALTinterwordspacing
H.~Sloetjes and P.~Wittenburg, ``Annotation by category-elan and iso dcr,'' in \emph{6th international Conference on Language Resources and Evaluation (LREC 2008)}, 2008. [Online]. Available: \url{https://archive.mpi.nl/tla/elan}
\BIBentrySTDinterwordspacing

\bibitem{baltruvsaitis2016openface}
T.~Baltru{\v{s}}aitis, P.~Robinson, and L.-P. Morency, ``Openface: an open source facial behavior analysis toolkit,'' in \emph{2016 IEEE Winter Conference on Applications of Computer Vision (WACV)}.\hskip 1em plus 0.5em minus 0.4em\relax IEEE, 2016, pp. 1--10.

\bibitem{zhang2015appearance}
X.~Zhang, Y.~Sugano, M.~Fritz, and A.~Bulling, ``Appearance-based gaze estimation in the wild,'' in \emph{Proceedings of the IEEE conference on computer vision and pattern recognition}, 2015, pp. 4511--4520.

\bibitem{macpherson2017attentional}
A.~C. MacPherson and C.~Moore, ``Attentional control by gaze cues in infancy,'' in \emph{Gaze-following: Its development and significance}.\hskip 1em plus 0.5em minus 0.4em\relax Psychology Press, 2017, pp. 53--75.

\bibitem{vernon2014fostering}
T.~W. Vernon, ``Fostering a social child with autism: A moment-by-moment sequential analysis of an early social engagement intervention,'' \emph{Journal of autism and developmental disorders}, vol.~44, pp. 3072--3082, 2014.

\bibitem{mundy1997joint}
P.~Mundy and M.~Crowson, ``Joint attention and early social communication: Implications for research on intervention with autism,'' \emph{Journal of Autism and Developmental disorders}, vol.~27, pp. 653--676, 1997.

\end{thebibliography}
\balance

\end{document}


\title{Gaze Behavior During a Long-Term, In-Home, Social Robot Intervention for Children with ASD\\
}
\author{
    Rebecca Ramnauth$^{1}$, Frederick Shic$^{2, 3}$, Brian Scassellati$^{1}$ \\
    \IEEEauthorblockA{
        \small
        $^{1}$Department of Computer Science, Yale University, New Haven, CT, USA \\
        $^{2}$Center for Child Health, Behavior, and Development, Seattle Children’s Research Institute, Seattle, WA, USA \\
        $^{3}$Department of Pediatrics, University of Washington School of Medicine, Seattle, WA, USA \\
        {\tt\small rebecca.ramnauth@yale.edu}
    }
}
\maketitle
\vspace{-25pt}

\section*{Appendix I}

\noindent \textbf{Children's Average Gaze Duration \& Frequency Per Week.} The change in children's average gaze duration (on the left) and gaze instances (on the right) with each attentional target per intervention week are shown. Overall, the intervention results in significant increases in the children's gaze and distribution of attention toward their caregivers as compared to other targets within and outside of the intervention. Additionally, we observe several week-dependent changes in behavioral variability and improvements in gaze behavior after the second week of the intervention.

\begin{figure}[H]  
  \centering  
  \includegraphics[width=0.95\linewidth]{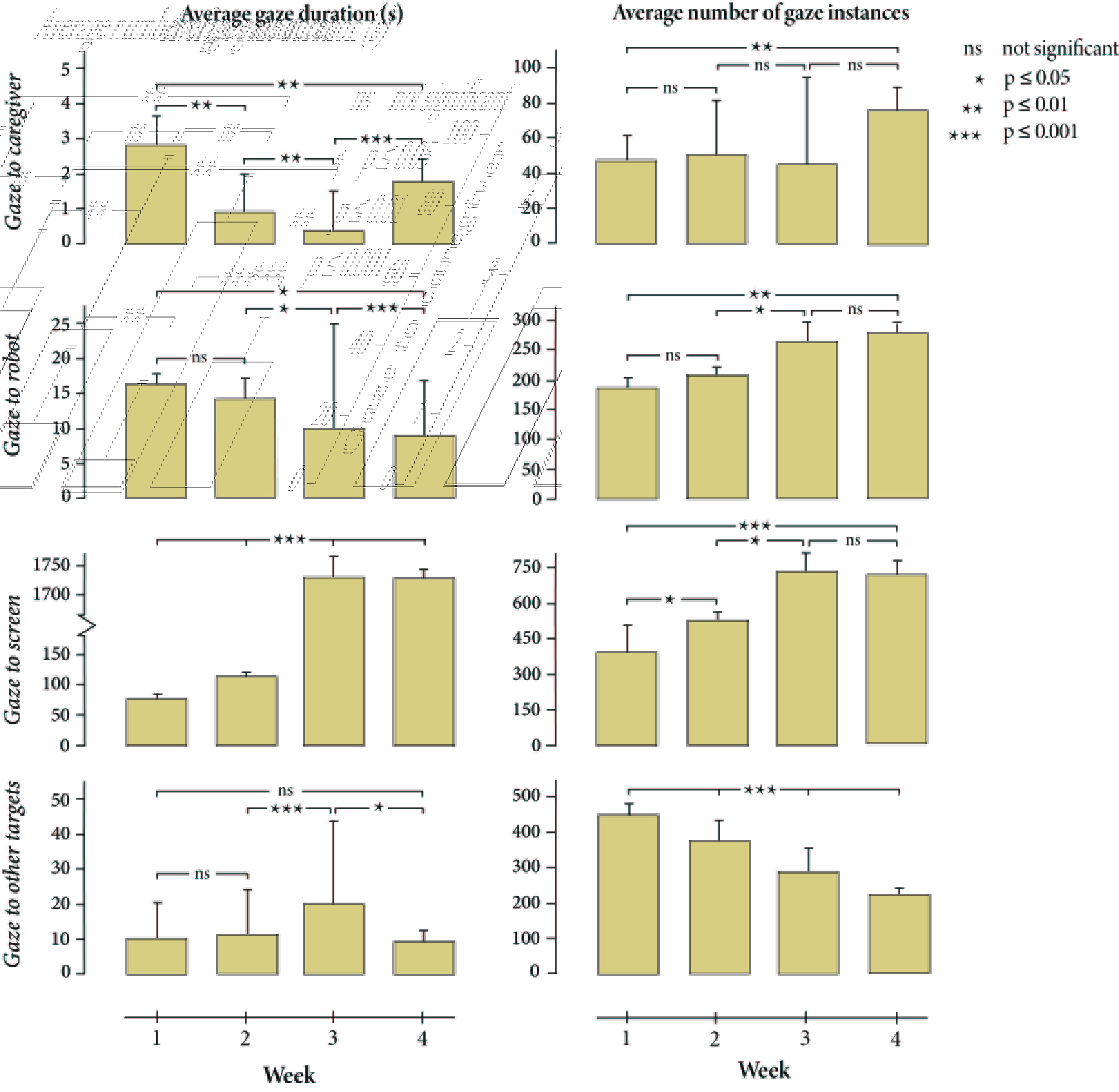}
  \label{fig:child}  
\end{figure}

\section*{Appendix II}

\noindent \textbf{Caregivers' Average Gaze Duration \& Frequency Per Week.} The change in caregivers' average gaze duration (on the left) and gaze instances (on the right) with each attentional target per intervention week are shown. 

\begin{figure}[H]  
  \centering  
  \includegraphics[width=0.95\linewidth]{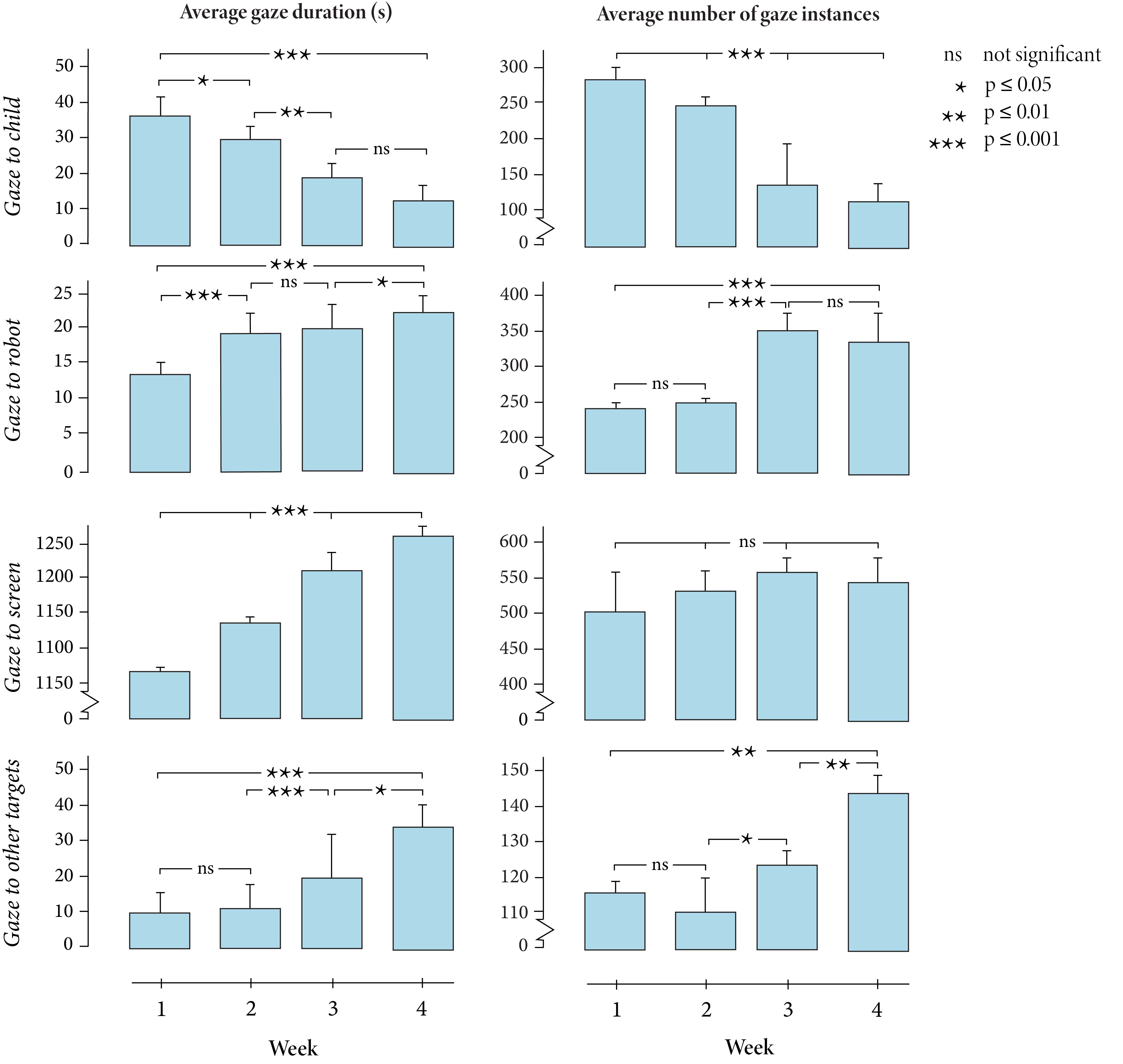}
  \label{fig:parent}  
\end{figure}

